%% file: main.tex
\documentclass[runningheads]{llncs}
\usepackage[T1]{fontenc}
\usepackage{graphicx}
\usepackage{booktabs}
\usepackage[misc]{ifsym}
\newcommand{\corr}{(\Letter)}
\usepackage{mwe}

\input{own_command}

\begin{document}

\title{
Efficiently Predicting Mutational Effect on Homologous Proteins by Evolution Encoding
}

\titlerunning{Evolution-aware Message Passing Neural Network}

\author{
Zhiqiang Zhong\inst{1} \corr
\and 
Davide Mottin\inst{1}
}

\authorrunning{Z. Zhong and D. Mottin}

\institute{
Aarhus University, Denmark \email{\{zzhong,davide\}@cs.au.dk}
}

\maketitle              

\begin{abstract} 
\input{pages/abstract}
\end{abstract}

\section{Introduction} 
\label{sec:introduction}
\input{pages/introduction}

\section{Preliminary and Problem} 
\label{sec:preliminary}
\input{pages/preliminary}

\section{Related Work} 
\label{sec:related_work}
\input{pages/related_work}

\section{Framework} 
\label{sec:framework}
\input{pages/framework}

\section{Experimental Study} 
\label{sec:experimental_study}
\input{pages/experiments}

\section{Conclusion and Future Work} 
\label{sec:conclusion}
\input{pages/conclusion}

\section*{Acknowledgments}
This work is supported by the Horizon Europe and Innovation Fund Denmark under the Eureka, Eurostar grant no E115712 - AAVanguard.

%
%
%
\bibliographystyle{splncs04}
\bibliography{full_format_references}





\newpage
\appendix

\input{pages/appendix}

\end{document}

%% file: own_command.tex
\usepackage{adjustbox}
\usepackage{makecell}
\usepackage{multirow}
\usepackage[ruled,vlined,linesnumbered]{algorithm2e}
\usepackage{enumitem}
\usepackage{amsmath}
\usepackage{amsfonts}
\usepackage{xcolor}
\newtheorem{assumption}{Assumption}
\usepackage{wrapfig}
\usepackage{algorithmic}
\usepackage{booktabs}
\usepackage{hyperref}
\usepackage{mdframed}
\newcommand{\specialcell}[2][c]{%
    \begin{tabular}[#1]{@{}l@{}}#2\end{tabular}
}

\newcommand{\AAV}{\texttt{AAV}\xspace}
\newcommand{\GB}{\texttt{GB1}\xspace}
\newcommand{\Fluorescence}{\texttt{Fluorescence}\xspace}
\newcommand{\wt}{wild-type\xspace}

\newcommand{\eg}{\emph{e.g.}}
\newcommand{\ie}{\emph{i.e.}}
\newcommand{\model}{\textsc{EvolMPNN}\xspace}
\newcommand{\gnn}{\textsc{EvolGNN}\xspace}
\newcommand{\former}{\textsc{EvolFormer}\xspace}


%% file: pages/abstract.tex
Predicting protein properties is paramount for biological and medical advancements. 
Current protein engineering mutates on a typical protein, called the \emph{\wt}, to construct a family of homologous proteins and study their properties. 
Yet, existing methods easily neglect subtle mutations, failing to capture the effect on the protein properties. 
To this end, we propose \model, \underline{Evol}ution-aware \underline{M}essage \underline{P}assing \underline{N}eural \underline{N}etwork, an efficient model to learn evolution-aware protein embeddings. 
\model samples sets of anchor proteins, computes evolutionary information by means of residues and employs a differentiable evolution-aware aggregation scheme over these sampled anchors.
This way, \model can efficiently utilise a novel message-passing method to capture the mutation effect on proteins with respect to the anchor proteins. 
Afterwards, the aggregated evolution-aware embeddings are integrated with sequence embeddings to generate final comprehensive protein embeddings. 
Our model shows up to $6.4\%$ better than state-of-the-art methods and attains $36\times$ inference speedup in comparison with large pre-trained models. 
Code and models are available at \url{https://github.com/zhiqiangzhongddu/EvolMPNN}.

%% file: pages/introduction.tex
\emph{Can we predict important properties of a protein by directly observing only the effect of a few mutations on such properties?} 
This basic biological question~\cite{FF14} has recently engaged the machine learning community due to the current availability of benchmark data~\cite{RBTDCCAS19,DMJWBGMY21,XZLZZMLT22}. 
Proteins are sequences of amino acids (or residues), which are the cornerstone of life and influence a number of metabolic processes, including diseases~\cite{PCB51}. 
For this reason, protein engineering stands at the forefront of modern biotechnology, offering a remarkable toolkit to manipulate and optimise existing proteins for a wide range of applications, from drug development to personalised therapy~\cite{AKBAC19}.

\input{figures/fig-intro}

One fundamental process in protein engineering progressively mutates an initial protein, called the \emph{\wt}, to study the effect on the protein's properties~\cite{SVLD91}. 
These mutations form a family of \emph{homologous proteins} as in Figure~\ref{fig:illustraton_research_question}. 
This process is appealing due to its cheaper cost compared to other methods and reduced time and risk~\cite{WSZ12}. 

Yet, the way mutations affect the protein's properties is not completely understood~\cite{BBSJORCCK21}, as it depends on a number of chemical reactions and bonds among residues. 
For this reason, machine learning offers a viable alternative to model complex interactions among residues. 
Initial approaches employed \emph{feature engineering} to capture protein's evolution~\cite{SG15,FZ00}; yet, a manual approach is expensive and does not offer enough versatility. 
Advances in NLP and CV inspired the design of deep \emph{protein sequence encoders}~\cite{HS97,VSPUJGKP17} and general purpose Protein Language Models (PLMs) that are pre-trained on large-scale datasets of sequences. 
Notable PLMs include ProtBert~\cite{BOPRL22}, AlphaFold~\cite{JEPGFRTBZP21}, TAPE Transformer~\cite{RBTDCCAS19} and ESM~\cite{RMSGLLGOZM21}. 
These models mainly rely on Multiple Sequence Alignments (MSAs)~\cite{MRVLSR21} to search on large databases of protein evolution. 
While this process focuses on conserved regions, it is insensitive to subtle yet crucial mutations in less conserved regions and introduces additional computational burdens~\cite{P13,CMCKBEN16}.

To overcome the limitations of previous models, we propose \model, \underline{Evol}ution-aware \underline{M}essage \underline{P}assing \underline{N}eural \underline{N}etwork, to predict the mutational effect on homologous proteins. 
Our fundamental assumption is that there are inherent correlations between protein properties and the sequence differences among them, as shown in Figure~\ref{fig:illustraton_research_question}-(b). 
\model devises a novel message-passing method to integrate both protein sequence and evolutionary information by identifying where and which mutations occur on the target protein sequence, compared with known protein sequences and predicts the mutational effect on the target protein property. 
To avoid the costly \emph{quadratic} pairwise comparison among proteins, we devise a theoretically grounded (see Section~\ref{subsec:theoretical_analysis}) \emph{linear} sampling strategy to compute differences only among the proteins and a fixed number of anchor proteins (Section~\ref{subsec:emb_update}). 
We additionally introduce two extensions of our model, \gnn and \former, to include available data on the relation among proteins (Section~\ref{subsec:extension}). 
The theoretical computation complexities of proposed methods are provided to guarantee their efficiency and practicality. 
We apply the proposed methods to three benchmark homologous protein property prediction datasets with nine splits. 
Empirical evaluation results (Section~\ref{subsec:experimental_results}) show up to $6.7\%$ Spearman's $\rho$ correlation improvement over the best performing baseline models, reducing the inference time by $36\times$ compared with pre-trained PLMs. 

%% file: figures/fig-intro.tex
\begin{figure}[!ht]
\centering
\includegraphics[width=1.\linewidth]{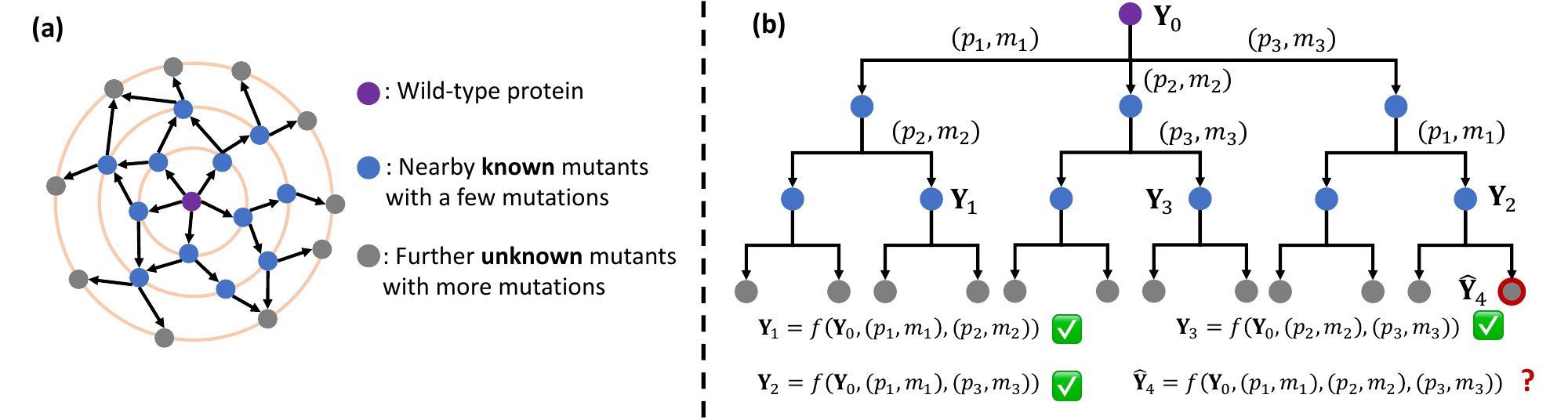}
\caption{
Protein property prediction on homologous protein family. 
(a) An example homologous protein family with labelled nearby mutants with few mutations. We aim to predict the label of unknown mutants with more mutations.
(b) The evolutionary pattern for (a).
For instance, $\mathbf{Y}_0$ is the label vector of the corresponding protein sequence, and $(p_1, m_1)$ indicates mutation $m_1$ at position $p_1$ of the protein's amino acid sequence. 
}
\label{fig:illustraton_research_question}
\end{figure}

%% file: pages/preliminary.tex
In protein engineering, we first receive a \emph{set of proteins} $\mathcal{M} = \{\mathcal{P}_{i}\}_{i = 1, 2, \dots, M}$ in which each protein can be associated with a label vector $\mathbf{Y}_i \in \mathbb{R}^{\theta}$ that describes its biomedical properties, \eg, fitness, stability, fluorescence, and solubility.
Each protein $\mathcal{P}_i$ is a linear chain of \emph{amino-acids} $\mathcal{P}_i = \{r_j\}_{j = 1, 2, \dots, N}$. 
While a protein sequence folds into specific 3D conformation to perform some biomedical functions, each amino-acid is considered as a \emph{residue}. 
Residues form peptide bonds and can interact with each other by different chemical bounds~\cite{PCB51}. 
In short, the function of a protein is mainly determined by the chemical interactions between residues. 
Since the 3D structure is missing in benchmark datasets~\cite{RBTDCCAS19,DMJWBGMY21,XZLZZMLT22}, we assume no 3D protein information in this paper. 

\smallskip\noindent
\textbf{Homologous Protein Family.}
A set of protein sequences ($\mathcal{M}$) is a \emph{homologous protein family} if there exists an ancestral protein $\mathcal{P}_{\textsc{WT}}$, called \emph{\wt}, such that any $\mathcal{P}_i \in \mathcal{M}$ is obtained by mutating $\mathcal{P}_{\textsc{WT}}$ through substitution, deletion, insertion and truncation of residues~\cite{OVSC19}. 
As shown in Figure~\ref{fig:illustraton_research_question}-(a), a homologous protein family can be organised together by representing their evolutionary relationships and Figure~\ref{fig:illustraton_research_question}-(b) illustrates the detailed evolutionary patterns. 

\smallskip\noindent
\textbf{Research Problem.}
Protein engineering based on homologous proteins is a promising and essential direction for designing novel proteins of desired properties~\cite{FF14}. 
Understanding the relation between protein sequence and property is one essential step. 
Practically, biologists perform experiments in the lab to manually label the property $\mathbf{Y}_{\textsc{Train}}$ of a set of protein $\mathcal{M}_{\textsc{Train}} \subset \mathcal{M}$ and the follow-up task is predicting $\hat{\mathbf{Y}}_{\textsc{Test}}$ of the rest proteins $\mathcal{M}_{\textsc{Test}} \subset \mathcal{M}$. 
However, due to their shared ancestry, homologous proteins typically have similarities in their amino acid sequences, structures, and functions.
\emph{Accurately predicting the homologous protein property by distinguishing these subtle yet crucial differences is still an open challenge.}

%% file: pages/related_work.tex
\textbf{Feature Engineering.}
Besides conducting manual experiments in labs to measure protein properties, the basic solution is to design different feature engineering methods based on relevant biological knowledge, to extract useful information from protein sequence~\cite{FZ00}. 
\cite{DMJWBGMY21} introduce using Levenshtein distance~\cite{LL07} and BLOSUM62-score~\cite{E04} relative to \wt to design protein sequence features. 
In another benchmark work, \cite{XZLZZMLT22} adopt another two typical protein sequence feature descriptors, \ie, Dipeptide Deviation from Expected Mean (DDE)~\cite{SG15} and Moran correlation (Moran)~\cite{FZ00}. 
For more engineering methods, refer to the comprehensive review~\cite{LRO07}. 

\smallskip\noindent
\textbf{Protein Representation Learning.}
In the last decades, propelled by the outstanding achievements of machine learning and deep learning, protein representation learning has revolutionised protein property prediction research. 
Early work along this line adopts the idea of word2vec~\cite{MSCCD13} to protein sequences. 
To increase model capacity, deeper \emph{protein sequence encoders} were proposed by the Computer Vision and Nature Language Processing communities~\cite{HS97,VSPUJGKP17}. 
The latest works develop \emph{Protein Language Models}, which focus on employing deep sequence encoder models for protein sequences and are pre-trained on million- or billion-scale sequences. 
Well-known works include ProtBert~\cite{BOPRL22}, AlphaFold~\cite{JEPGFRTBZP21}, TAPE Transformer~\cite{RBTDCCAS19}, ESM~\cite{RMSGLLGOZM21}, AlphaFold2~\cite{JEPGFRTBZP21} and Ankh~\cite{EESMERR23}. 
However, most current research on protein sequences defines it as a language problem and does not sufficiently consider the subtle yet crucial evolutionary patterns in homologous proteins.
For instance, \cite{RMSGLLGOZM21,JEPGFRTBZP21} explore protein Multiple Sequence Alignments (MSAs)~\cite{RLVMCASR21,MRVLSR21} to capture the mutational effect.
Nevertheless, the MSA searching process introduces additional computational burden and is insensitive to subtle but crucial sequence differences~\cite{P13}. 
\cite{CMCKBEN16} indicate the shortcomings of MSAs on easily neglecting the presence of minor mutations, which can propagate errors to downstream protein sequence representation learning tasks. 
The message-passing idea has been explored in protein folding tasks~\cite{JEPGFRTBZP21,DABB22}. 
We build on this direction, by devising a novel model that captures subtleties in the mutations for homologous protein property prediction tasks. 

%% file: pages/framework.tex
\model is a novel framework that integrates both protein sequence information and evolution information by means of residues. 
As a result, \model accurately predicts the mutational effect on homologous protein families.
First, in Section~\ref{subsec:emb_init}, we introduce \emph{embedding initialisation} for protein sequence and residues and the update module for residue embedding (Section~\ref{subsec:emb_update}).
The \emph{evolution encoding} in Section~\ref{subsec:evolution_encode} is the cornerstone of the model that ameliorates protein embeddings. We conclude in Section~\ref{subsec:final_emb} with the generation of \emph{final proteins embeddings and model optimisation}. 
We complement our model with a theoretical analysis to motivate our methodology and a discussion of the computation complexity (Section~\ref{subsec:theoretical_analysis}). 
We additionally propose extended versions of \model that deal with available protein-protein interactions (Section~\ref{subsec:extension}).

\input{figures/fig-framework}

\subsection{Embedding Initialisation}
\label{subsec:emb_init}

\textbf{Protein Sequence Embedding.}
Given a \emph{set of proteins} $\mathcal{M} = \{\mathcal{P}_{i}\}_{i = 1, 2, \dots, M}$, 
we first adopt a (parameter-frozen) PLM model~\cite{RMSOR21,MRVLSR21}\footnote{We do not fine-tune PLM in this paper for efficiency consideration.} as protein sequence encoder to initialise protein-sequence embeddings ($\mathbf{H}$) as one $d$-dimensional real-valued vector for every protein $\mathcal{P}_i$, which include \emph{macro} (\ie, protein sequence) level information as the primary embedding.
\begin{equation}
\label{eq:protein_emb}
    \mathbf{H} = \textsc{PlmEncoder}(\{ \mathcal{P}_{i} \}_{i = 1, 2, \dots, M}),
\end{equation}
where the obtained protein embedding $\mathbf{H} \in \mathbb{R}^{M \times d}$ and $\mathbf{H}_i$ corresponds to each protein $\mathcal{P}_i$. 
Different encoders can extract information on various aspects, however, existing PLM models that rely on MSAs are not sensitive enough to capture the evolution pattern information in homologous protein families~\cite{P13}. 
\cite{CMCKBEN16} systematically indicate the shortcomings of MSAs on easily neglecting the presence of minor mutations, which can propagate errors to downstream protein sequence representation learning tasks. 

\smallskip\noindent
\textbf{Residue Embedding Initialisation.}
In order to properly capture the evolution information in homologous proteins, we delve into the residue level for \emph{micro} clues. 
We adopt two residue embedding initialisation approaches, \ie, one-hot encoding ($\Phi^{\textsc{OH}}$) and pre-trained PLM encoder ($\Phi^{\textsc{PLM}}$), to generate protein's initial residue embeddings $\mathbf{X}_i = \{ \mathbf{x}^i_j \}_{j = 1, 2, \dots, N}$, where $\mathbf{x}^i_j \in \mathbb{R}^{d}$. 
In particular, $\Phi^{\textsc{OH}}$ assigns each protein residue\footnote{There are $20$ different amino acid residues commonly found in proteins} with a binary feature vector $\mathbf{x}^i_{j}$, where $\mathbf{x}^i_{jb} = 1$ indicates the appearance of the $b$-th residue at $\mathcal{P}_i$'s $j$-th position.
By stacking $N$ residues' feature vectors into a matrix, we can obtain $\mathbf{X}_i \in \mathbb{R}^{N \times d}$. 
On the other hand, following the benchmark implementations~\cite{ZSZLXYZCCLMLXQT22}, \textsc{PlmEncoder} can export residue embeddings similar to Eq.~\ref{eq:protein_emb}. 
Formally, $\Phi^{\textsc{PLM}}$ initialises protein residue embeddings as $\mathbf{X}_i = \textsc{PlmEncoder}(\{ r_{j} \}_{j = 1, 2, \dots, N})$.

\smallskip\noindent
\textbf{Position Embedding.}
Another essential component of existing PLM is the positional encoding, which was first proposed by~\cite{VSPUJGKP17}.
This positional encoding effectively captures the relative structural information between residues and integrates it with the model. 
In our case, correctly recording the position of each residue in the protein sequence plays an essential role in identifying each protein's corresponding mutations. 
Because a mutation occurring at different positions can have varying effects on protein properties due to its impact on protein structure.
Therefore, after initialising residue embeddings, we further apply positional embedding on each protein's residues.
Particularly, we randomly initialise a set of $d$ position embeddings $\Phi^{\textsc{Pos}} \in \mathbb{R}^{N \times d}$, and it will be learned in the training process of the entire framework. 
We denote the residue embedding empowered by position embedding as $\hat{\mathbf{X}}_i = \mathbf{X}_i \odot \Phi^{\textsc{Pos}}$, where $\odot$ indicates the element-wise multiplication. 

\subsection{Residue Embedding Update}
\label{subsec:emb_update}

To maintain a stable 3D structure, 3D protein folding depends on the strength of different chemical bonds between residues.
Previous studies manually designed residue contact maps to model the residue-residue interactions to learn effective residue embeddings~\cite{RMSOR21,GJZJLZYHL23}. 
In this paper, we adopt the residue-residue interaction to update residue embeddings but 
eschew the requirement of manually designing the contact map. 
Instead, we assume the existence of an implicit fully connected residue contact map of each protein $\mathcal{P}_i$ and implement the Transformer model~\cite{VSPUJGKP17,WZLWY22} to adaptively update residue embeddings. 
Denote $\mathbf{R}^{(\ell)}_i$ as the input to the $(\ell + 1)$-th layer, with the first $\mathbf{R}^{(0)}_i = \hat{\mathbf{X}}_i$ be the input encoding. 
The $(\ell + 1)$-th layer of residue embedding update module can be formally defined as follows:
\begin{equation}
\label{eq:transformer}
\begin{aligned}
    \mathbf{Att}^h_i (\mathbf{R}^{(\ell)}_i) &= \textsc{Softmax}(\frac{\mathbf{R}^{(\ell)}_i \mathbf{W}^{\ell, h}_Q (\mathbf{R}^{(\ell)}_i \mathbf{W}^{\ell, h}_K)^{\textsc{T}} }{\sqrt{d}}), \\
    \hat{\mathbf{R}}^{(\ell)}_i &= \mathbf{R}^{(\ell)}_i + \sum^{H}_{h=1} \mathbf{Att}^h_i (\mathbf{R}^{(\ell)}_i) \mathbf{R}^{(\ell)}_i \mathbf{W}^{\ell, h}_{V} \mathbf{W}^{\ell, h}_{O}, \\
    \mathbf{R}^{(\ell + 1)}_i &= \hat{\mathbf{R}}^{(\ell)}_i + \textsc{ELU}(\hat{\mathbf{R}}^{(\ell)}_i \mathbf{W}^{\ell}_1 )\mathbf{W}^{\ell}_{2},
\end{aligned}
\end{equation}
where $\mathbf{W}^{\ell, h}_{O} \in \mathbb{R}^{d_H \times d}$, $\mathbf{W}^{l, h}_Q$, $\mathbf{W}^{l, h}_K$, $\mathbf{W}^{l, h}_V \in \mathbb{R}^{d \times d_H}$, $\mathbf{W}^{\ell}_1 \in \mathbb{R}^{d \times r}$, $\mathbf{W}^{\ell}_2 \in \mathbb{R}^{d_t \times d}$ are learnable parameters, $H$ is the number of attention heads, $d_H$ is the dimension of each head, $d_t$ is the dimension of the hidden layer, 
\textsc{ELU}~\cite{CUH15} is an activation function, and $\mathbf{Att}^h_i (\mathbf{R}^{(\ell)}_i)$ refers to as the attention matrix. 
After each Transformer layer, we add a normalisation layer \ie, LayerNorm~\cite{BKH16}, to reduce the over-fitting problem proposed by~\cite{VSPUJGKP17}. 
After stacking $L_r$ layers, we obtain the final residue embeddings as $\mathbf{R}_i = \mathbf{R}^{(L_r)}_i$.

\subsection{Evolution Encoding}
\label{subsec:evolution_encode}

In homologous protein families, all proteins are mutants derived from a common \wt protein $\mathcal{P}_{\textsc{WT}}$ with different numbers and types of mutations. 
In this paper, we propose to capture the evolutionary information via the following assumption.
\begin{assumption}[Protein Property Relevance]
\label{assump:relevance}
    Assume there is a homologous protein family $\mathcal{M}$ and a function $\mathrm{F}_{\textsc{Diff}}$ can accurately distinguish the mutations on mutant $\mathcal{P}_i$ compared with any $\mathcal{P}_j$ as $\mathrm{F}_{\textsc{Diff}}(\mathcal{P}_i, \mathcal{P}_j)$. 
    For any target protein $\mathcal{P}_i$, its property $\mathbf{Y}_i$ can be predicted by considering 1) its sequence information $\mathcal{P}_i$; 2) $\mathrm{F}_{\textsc{Diff}}(\mathcal{P}_i, \mathcal{P}_j)$ and the property of $\mathcal{P}_j$, \ie, $\mathbf{Y}_j$. 
    Shortly, we assume there exists a function $f$ that maps $\mathbf{Y}_i \leftarrow f(\mathrm{F}_{\textsc{Diff}}(\mathcal{P}_i, \mathcal{P}_j), \mathbf{Y}_j)$. 
\end{assumption}

Motivated by Assumption~\ref{assump:relevance}, we take both protein sequence and the mutants difference $\mathrm{F}_{\textsc{Diff}}(\mathcal{P}_i, \mathcal{P}_j)$ to accurately predict the protein property.
To encode the protein sequence, we employ established tools described in Section~\ref{subsec:emb_init}. 
Here instead, we describe the evolution encoding to realise the function of $\mathrm{F}_{\textsc{Diff}}(\mathcal{P}_i, \mathcal{P}_j)$. 

The na\"ive solution to extract evolutionary patterns in a homologous family is constructing a complete phylogenetic tree~\cite{FM67} based on the mutation distance between each protein pair. 
Yet, finding the most parsimonious phylogenetic tree is $\mathbf{NP}$-hard~\cite{S75}.

To address the aforementioned problems, instead of constructing the phylogenetic tree, we compute the distance among a few sampled proteins we call \emph{anchor proteins} and all the other proteins. 
Theoretical analysis to validate this design is discussed in Section~\ref{subsec:theoretical_analysis}. 
Specifically, denote $\mathbf{H}^{(\ell)}_i$ as the input to the $(\ell + 1)$-th block and define $\mathbf{H}^{(0)}_i = \mathbf{H}_i$. 
The evolution localisation encoding  of the $(\ell + 1)$-th layer contains the following key components:
\textit{(i)} $k$ anchor protein $\{ \mathcal{P}_{S_i} \}_{i = 1, 2, \dots, k}$ selection.
\textit{(ii)} Evolutionary information encoding function $\mathrm{F}_{\textsc{Diff}}$ that computes the difference between residues of each protein and those of the anchor protein, and target protein's evolutionary information is generated by summarising the obtained differences $\mathbf{d}_{ij}$ as follows:
\begin{equation}
\label{eq:combine_diff}
    \mathbf{d}_{ij} = \textsc{Combine}(\mathbf{R}_i - \mathbf{R}_{S_j}), \\
\end{equation}
where $\textsc{Combine}$ can be implemented as differentiable operators, such as, \textsc{Concatenate}, \textsc{MAX Pool} \textsc{MEAN Pool} and \textsc{SUM Pool}; here we use the \textsc{MEAN Pool} to obtain $\mathbf{d}_{ij} \in \mathbb{R}^{d}$.
\textit{(iii)} Message computation function $\mathrm{F}_{\textsc{Message}}$ that integrates protein features and evolutionary information as one message from an anchor protein. 
Specifically, $\mathrm{F}_{\textsc{Message}}$ combines protein sequence feature information of two proteins with their evolutionary differences. 
We empirically find that the simple element-wise product between $\mathbf{H}^{(\ell)}_j$ and $\mathbf{d}_{ij}$ attains good results
\begin{equation}
    \mathrm{F}_{\textsc{Message}}(i, j, \mathbf{H}^{(\ell)}_j, \mathbf{d}_{ij}) = \mathbf{H}^{(\ell)}_j \odot \mathbf{d}_{ij},
\end{equation}
\textit{(iv)} Aggregate evolutionary messages from $k$ anchors and combine them with the protein's embedding as the updated protein embedding, which contains the protein sequence and evolutionary information:
\begin{gather}
    \label{eq:combine}
    \hat{\mathbf{H}}^{(\ell)}_i = \textsc{Combine}(\{\mathrm{F}_{\textsc{Message}}(i, j, \mathbf{H}^{(\ell)}_j, \mathbf{d}_{ij} ) \}_{j = 1, 2, \dots, k} ), \\
    \label{eq:concat}
    \mathbf{H}^{(\ell + 1)}_i = \textsc{Concat}(\mathbf{H}^{(\ell)}_i, \hat{\mathbf{H}}^{(\ell)}_i) \mathbf{W}^{\ell},
\end{gather}
where $\mathbf{W}^{\ell} \in \mathbb{R}^{2d \times d}$ transform concatenated vectors to the hidden dimension. 
After stacking $L_p$ layers, we obtain the final protein sequence embedding $\mathbf{Z}^{\textsc{P}}_i = \mathbf{H}^{(L_p)}_i$. 

\subsection{Final Embedding and Optimisation}
\label{subsec:final_emb}

After obtaining protein $\mathcal{P}_i$'s residue embeddings $\mathbf{R}_i$ and sequence embedding $\mathbf{Z}^{\textsc{P}}_i$, we summarise its residue embeddings as a vector $\mathbf{Z}^{\textsc{R}}_i = \textsc{MEAN Pool}(\mathbf{R}_i)$. 
The final protein embedding summarises the protein sequence information and evolution information as the comprehensive embedding $\mathbf{Z}_i = \textsc{Concat}(\mathbf{Z}^{\textsc{P}}_i, \mathbf{Z}^{\textsc{R}}_i)$ and the final prediction is computed as $\hat{\mathbf{Y}}_i = \mathbf{Z}_i \mathbf{W}^{\textsc{Final}}$ where $\mathbf{W}^{\textsc{Final}} \in \mathbb{R}^{d \times \theta}$, $\theta$ is the number of properties to predict. 
Afterwards, we adopt a simple and common strategy, similar to~\cite{XZLZZMLT22}, to solve the protein property prediction tasks. 
Specifically, we adopt the MSELoss ($\mathcal{L}$) to measure the correctness of model predictions on training samples against ground truth labels. 
The objective of learning the target task is to optimise model parameters to minimise the loss $\mathcal{L}$ on this task. 
The framework of \model is summarised in Algorithm~\ref{alg:framework}. 

\begin{algorithm}[!ht]
\caption{The framework of \model}
\label{alg:framework}
    \SetAlgoLined
    \KwIn{
        Protein set $\mathcal{M} = \{\mathcal{P}_{i}\}_{i = 1, 2, \dots, M}$ and each protein sequence $\mathcal{P}_i$ contains a residue set $\{r_j\}_{j = 1, 2, \dots, N}$;
        Message computation function $\mathrm{F}_{\textsc{Message}}$ that outputs an $d$ dimensional message; 
        $\textsc{Combine}(\cdot)$ and $\textsc{Concat}(\cdot)$ operators. 
    }
    \KwOut{
        Protein embeddings $\{ \mathbf{Z}_i \}_{i=1, 2, \dots, M}$
    }
    $\mathbf{H}_i \leftarrow \textsc{PlmEncoder}( \mathcal{P}_{i})$ \\
    $\mathbf{X}_i \leftarrow \Phi^{\textsc{OH}}(\{ r_{j} \}_{j = 1, 2, \dots, N}) \;/\; \Phi^{\textsc{PLM}}(\{ r_{j} \}_{j = 1, 2, \dots, N})$ \\
    $\hat{\mathbf{X}}_i \leftarrow \mathbf{X}_i \odot \Phi^{\textsc{Pos}}$ \\
    $ \mathbf{R}^{(0)}_i \leftarrow \hat{\mathbf{X}}_i$ \\
    \For{$\ell = 1, 2, \dots, L_r$}{
        \For{i = 1, 2, \dots, N}{
            $\mathbf{R}^{(\ell)}_i \leftarrow \textsc{NodeFormer}(\mathbf{R}^{(\ell-1)}_i)$ 
        }
    }
    $\mathbf{R}_i \leftarrow \mathbf{R}^{(L_r)}_i$ \\
    $\mathbf{H}^{(0)}_i \leftarrow \mathbf{H}_i$ \\
    \For{$\ell = 1, 2, \dots, L_p$}{
        $\{ S_j \}_{j = 1, 2, \dots, k} \sim \mathcal{M} $ \\
        \For{ $i = 1, 2, \dots, M$ }{
            \For{j = 1, 2, \dots, k}{
                $\mathbf{d}_{ij} = \textsc{Combine}(\mathbf{R}_i - \mathbf{R}_{S_j})$
            }
            $\hat{\mathbf{H}}^{(\ell)}_i = \textsc{Combine} ( \{\mathrm{F}_{\textsc{Message}}(i, j, \mathbf{H}^{(\ell)}_j, \mathbf{d}_{ij} ) \}_{j = 1, 2, \dots, k} $) 
            $\mathbf{H}^{(\ell + 1)}_i = \textsc{Concat}(\mathbf{H}^{(\ell)}_i, \hat{\mathbf{H}}^{(\ell)}_i) \mathbf{W}^{\ell}$ 
        }
    }
    $\mathbf{Z}^{\textsc{P}}_i = \mathbf{H}^{(L_p)}_i$ \\
    $\mathbf{Z}^{\textsc{R}}_i = \textsc{MEAN Pool}(\mathbf{R}_i)$ \\
    $\mathbf{Z}_i = \textsc{Concat}(\mathbf{Z}^{\textsc{P}}_i, \mathbf{Z}^{\textsc{R}}_i)$
\end{algorithm}

\subsection{Extensions on Observed Graph}
\label{subsec:extension}

\model does not leverage any information from explicit geometry among proteins, where each protein only communicates with randomly sampled anchors (Section~\ref{subsec:evolution_encode}). 
However, it is often possible to have useful structured data $G = (\mathcal{M}, \mathbf{A})$ that represents the relation between protein-protein by incorporating specific domain knowledge~\cite{ZBM23}.\footnote{Available contact map describes residue-residue interactions can be easily integrated as relational bias of Transformer~\cite{WZLWY22} as we used in Section~\ref{subsec:emb_update}.}
Therefore, here we introduce \gnn, an extension of \model on the possibly observed protein interactions. 

\smallskip\noindent
\textbf{\gnn.}
We compute the evolution information as Eq.~\ref{eq:combine_diff}. 
The evolution information can be easily integrated into the pipeline of message-passing neural networks, as an additional structural coefficient:
\begin{equation}
\label{eq:mpnn}
\begin{aligned}
\mathbf{m}^{(\ell)}_a &= \textsc{Aggregate}^{\mathcal{N}}(\{\mathbf{A}_{ij}, \underbrace{ \textcolor{orange}{\mathbf{d}_{ij}} }_{\textcolor{orange}{\text{Evol. info.}}}, \mathbf{H}^{(\ell-1)}_j \, \vert \, j \in \mathcal{N}(i) \}), \\
\mathbf{m}^{(\ell)}_i &= \textsc{Aggregate}^{\mathcal{I}}(\{\mathbf{A}_{ij}, \underbrace{ \textcolor{orange}{\mathbf{d}_{ij}} }_{\textcolor{orange}{\text{Evol. info.}}} \, \vert \, j \in \mathcal{N}(i) \}) \, \mathbf{H}^{(\ell-1)}_i, \\
\mathbf{H}^{(\ell)}_i &= \textsc{Combine}(\mathbf{m}^{(\ell)}_a, \mathbf{m}^{(\ell)}_i),
\end{aligned}
\end{equation}
where $\textsc{Aggregate}^{\mathcal{N}}(\cdot)$ and $\textsc{Aggregate}^{\mathcal{I}}(\cdot)$ are two parameterised functions. 
$\mathbf{m}^{(\ell)}_a$ is a message aggregated from the neighbours $\mathcal{N}(i)$ of protein $\mathcal{P}_i$ and their structure ($\mathbf{A}_{ij}$) and evolution ($\mathbf{d}_{ij}$) coefficients. 
$\mathbf{m}^{(\ell)}_{i}$ is an updated message from protein $\mathcal{P}_i$ after performing an element-wise multiplication between $\textsc{Aggregate}^{\mathcal{I}}(\cdot)$ and $\mathbf{H}^{(\ell-1)}_i$ to account for structural and evolution effects from its neighbours. 
After, $\mathbf{m}^{(\ell)}_a$ and $\mathbf{m}^{(\ell)}_i$ are combined together to obtain the update embedding $\mathbf{H}^{(\ell)}_i$. 

\smallskip\noindent
\textbf{\former.}
Another extension relies on pure Transformer structure, which means the evolution information of $\mathcal{M}$ can be captured by every protein. 
The evolution information can be integrated into the pipeline of Transformer, as additional information to compute the attention matrix:
\begin{equation}
    \mathbf{Att}^h (\mathbf{H}^{(\ell)}) = \textsc{Softmax}(\frac{\mathbf{H}^{(\ell)} \mathbf{W}^{\ell, h}_Q (\mathbf{H}^{(\ell)} \mathbf{W}^{\ell, h}_K)^{\textsc{T}} }{\sqrt{d}} + \underbrace{\textcolor{blue}{\textsc{MEAN Pool}(\{ \mathbf{R}_i \}_{i=1, 2, \dots, M}) } }_{\textcolor{blue}{\text{Evol. info.}}} ),
\end{equation}
Other follow-up information aggregation and feature vector update operations are the same as the basic Transformer pipeline, as described in Eq.~\ref{eq:transformer}. 

\subsection{Theoretical Analysis}
\label{subsec:theoretical_analysis}

\textbf{Anchor Selection.}
Inspired by~\cite{YYL19}, we adopt Bourgain's Theorem~\cite{B85} to guide the random anchor number ($k$) of the evolution encoding layer.
Briefly, support by a constructive proof (Theorem~\ref{th:constructive_proof}~\cite{LLR95}) of Bourgain Theorem (Theorem~\ref{th:bourgain}), only $k = O(\log^2 M)$ anchors are needed to ensure the resulting embeddings are guaranteed to have low distortion (Definition~\ref{def:destortion}), in a given metric space $(\mathcal{M}, \mathrm{F}_{\textsc{Dist}})$. 
\model can be viewed as a generalisation of the embedding method of Theorem~\ref{th:constructive_proof}, where $\mathrm{F}_{\textsc{Dist}} (\cdot)$ is generalised via message passing functions (Eq~\ref{eq:combine_diff}-Eq.~\ref{eq:concat}). 
Therefore, Theorem~\ref{th:constructive_proof} offers a theoretical guide that $O(\log^2M)$ anchors are needed to guarantee low distortion embedding. 
Following this principle, \model choose $k = \log^2M$ random anchors, denoted as $\{ S_{j} \}_{j = 1, 2, \dots, \log^2 M}$, and we sample each protein in $\mathcal{M}$ independently with probability $\frac{1}{2^j}$. 
Detailed discussion and proof refer to Appendix~\ref{sec:appendix_theoretical_analysis}. 

\smallskip\noindent
\textbf{Complexity Analysis.}
The computation costs of \model, \gnn, and \former come from residue encoding and evolution encoding, since the protein sequence and residue feature initialisation have no trainable parameters. 
The residue encoder introduces the complexity of $O(MN)$ following an efficient implementation of NodeFormer~\cite{WZLWY22}. 
In the evolution encoding, \model performs communication between each protein and $\log^2 M$ anchors, which introduces the complexity of $O(M\log^2 M)$;
\gnn performs communication between each protein and $K$ neighbours with $O(KM)$ complexity;
\former performs communication between all protein pairs, which introduces the complexity of $O(M)$, following the efficient implement, NodeFormer. 
In the end, we obtain the total computation complexity of \model\xspace - $O((N + \log^2 M)M)$, \gnn\xspace - $O((N + K)M)$ and \former\xspace - $O((N + 1)M)$. 

%% file: figures/fig-framework.tex
\begin{figure}[!ht]
\centering
\includegraphics[width=1.\linewidth]{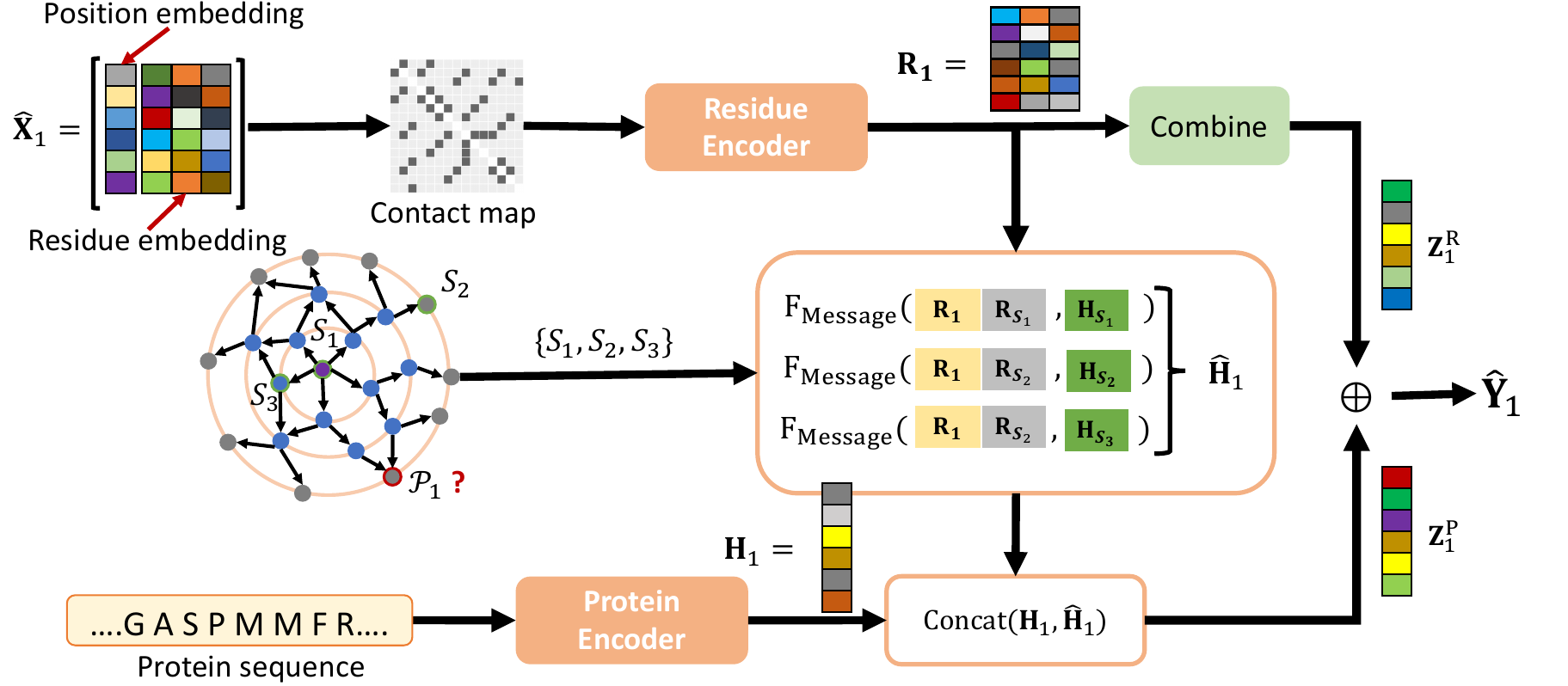}
\caption{
Our \model framework encodes protein mutations via a sapient combination of residue evolution and sequence encoding.
}
\label{fig:framework}
\end{figure}

%% file: pages/experiments.tex
In this section, we empirically study the performance of \model.
We validate our model on three benchmark homologous protein family datasets and evaluate the methods on nine data splits to consider comprehensive practical use cases. 
Our experiments comprise a comprehensive set of state-of-the-art methods from different categories. 
We additionally demonstrate the effectiveness of two extensions of our model, \gnn and \former, with different input features. 
We conclude our analysis by studying the influence of some hyper-parameters and investigating the performance of \model on high mutational mutants. 

\input{tables/tbl-dataset}

\smallskip\noindent
\textbf{Datasets and Splits.}
We perform experiments on benchmark datasets of several important protein engineering tasks, including \AAV, \GB and \Fluorescence, and generate three splits on each dataset. 
Data statistics are summarised in Table~\ref{table:landscapes_and_split_statistics}. 
The split \textsc{$\lambda$-vs-Rest} indicates that we train models on \wt protein and mutants of no more than $\lambda$ mutations, while the rest are assigned to test. 
The split \textsc{Low-vs-High} indicates that we train models on sequences with target value scores equal to or below \wt, while the rest are assigned to the test. 

\smallskip\noindent
\textbf{Baselines.}
As baseline models, we consider methods in four categories. 
First, we selected four \emph{feature engineer} methods, \ie, Levenshtein~\cite{DMJWBGMY21}, BLOSUM62~\cite{DMJWBGMY21}, DDE~\cite{SG15} and Moran~\cite{FZ00}. 
Second, we select four \emph{protein sequence encoder} models, \ie, LSTM~\cite{HS97}, Transformer~\cite{RBTDCCAS19}, CNN~\cite{RBTDCCAS19} and ResNet~\cite{YKF17}. 
Third, we select four \emph{pre-trained PLM models}, \ie, ProtBert~\cite{EHDRWJGFASBR22}, ESM-1b~\cite{RMSGLLGOZM21}, ESM-1v~\cite{MRVLSR21} and ESM-2~\cite{LARHZLSVKS23}.
In the end, we select four \emph{GNN-based methods} which can utilise available graph structure, \ie, GCN~\cite{KW17}, GAT~\cite{VCCRLB18}, GraphTransformer~\cite{SHFZWS21} and NodeFormer~\cite{WZLWY22}. 

\smallskip\noindent
\textbf{Implementation.}
We follow the PEER benchmark settings\footnote{\url{https://github.com/DeepGraphLearning/PEER_Benchmark}}, including train and test pipeline, model optimisation and evaluation method (evaluation is Spearman's $\rho$ metric), adopted in~\cite{XZLZZMLT22} to make sure the comparison fairness. 
For the baselines, including feature engineer, protein sequence encoder and pre-trained PLM, we adopt the implementation provided by benchmark Torchdrug~\cite{ZSZLXYZCCLMLXQT22} and the configurations reported in \cite{XZLZZMLT22}.  
For the GNN-based baselines, which require predefined graph structure and protein features, we construct $K$-NN graphs~\cite{EPY97}, with $K = \{5, 10, 15\}$, and report the best performance. As features, we use the trained sequence encoder, which achieves better performance, used also in our method. 
In addition, we adopt ESM-1b as the residue encoder on \GB dataset and adopt One-Hot encoding on \AAV and \Fluorescence datasets to speed up the training process. 
All experiments are conducted on two NVIDIA GeForce RTX 3090 GPUs and two NVIDIA RTX A6000 GPUs, and we report the mean performance of three runs with different random seeds. 
We present more details at \url{https://github.com/zhiqiangzhongddu/EvolMPNN}.

\subsection{Effectiveness}
\label{subsec:experimental_results}

\textbf{\model outperforms all baselines on 9 splits.}
Table~\ref{table:results} summarises performance comparison on \AAV, \GB and \Fluorescence datasets. 
\model achieves new state-of-the-art performance on most splits of three datasets, with up to $6.7\%$ improvements to baseline methods. 
This result vindicates the effectiveness of our proposed design to capture evolution information for homologous protein property prediction. 

\input{tables/tbl-results}

\smallskip\noindent
\textbf{Manual construction of homology graphs proves to be less effective.}
Notably, GNN-based methods that utilise manually constructed graph structure do not enter top-2 on 8 of 9 splits and two Transformer structure models, \ie, GraphTransformer and NodeFormer, often outperform such methods. 
It can be understood since homology graph construction is a challenging biomedical task~\cite{P13}, the simple $K$-NN graph construction is not an effective solution. 

\smallskip\noindent
\textbf{Large-scale PLM models are dominated by simple models.}
Surprisingly, we find that smaller models, such as CNN and ResNet, can outperform large ESM variants pre-trained on million- and billion-scale sequences.
For instance, ESM-1v has about 650 million parameters and is pre-trained on around 138 million UniRef90 sequences~\cite{MRVLSR21}. 
Yet, CNN outperforms ESM-1v on three splits of \Fluorescence dataset. 
This indicates the necessity of designs targeting specifically the crucial homologous protein engineering task. 

\input{tables/tbl-extensions}

\smallskip\noindent
\textbf{Our proposed extension models outperform all baselines on \GB dataset.}
We performed additional experiments on \GB datasets to investigate the performance of two extended models, \ie, \gnn and \former and study the influence of different residue embedding initialisation methods.
The results summarised in Table~\ref{table:results_gb1_variants} evince that \model outperforms the other two variants in three splits, and all our proposed models outperform the best baseline. 
This result confirms the effectiveness of encoding evolution information for homologous protein property prediction. 
Besides, the models adopting the PLM encoding $\Phi^{\textsc{PLM}}$ achieve better performance than those using the one-hot encoding $\Phi^{\textsc{OH}}$. 
From this experiment, we conclude that residue information provided by PLM helps to capture protein's evolution information. 

\subsection{Analysis of Performance}

\input{figures/fig-num-mutation}

\smallskip\noindent
\textbf{The performance of \model comes from its superior predictions on high mutational mutants.}
For the \textsc{Low-vs-High} split of \GB dataset, we group the test proteins into 4 groups depending on their number of mutations. 
Next, we compute three models, including \model, ESM-1b (fine-tuned PLM model) and CNN (best baseline), prediction performances on each protein group and present the results in Figure~\ref{fig:num_mutations}. 
\model outperforms two baselines in all 4 protein groups. 
Notably, by demonstrating \model's clear edge in groups of no less than 3 mutations, we confirm the generalisation effectiveness from low mutational mutants to high mutational mutants. 
\textbf{As per inference time}, \model and CNN require similar inference time ($\approx5$ secs), $36\times$ faster than ESM-1b ($\approx3$ mins).

\input{figures/fig-hyper-param}

\smallskip\noindent
\textbf{Influence of hyper-parameter settings on \model.}
We present in Figure~\ref{fig:hyper_param} a group of experiments to study the influence of some hyper-parameters on \model, including the number of attention heads, embedding dimension and the number of layers of residue encoder and evolution encoder. 

%% file: tables/tbl-dataset.tex
\begin{table}[!ht]
\caption{
    Dataset splits and corresponding statistics; if the split comes from a benchmark paper, we report the corresponding citation.
}
\label{table:landscapes_and_split_statistics}
\centering
\begin{tabular}{l|l|l|l|l|l}
\hline
\hline
\textbf{Landscape} & \textbf{Split} & \textbf{\# Total} & \textbf{\#Train} & \textbf{\#Valid} & \textbf{\#Test} \\
\hline
AAV~\cite{BBSJORCCK21} 
& \textsc{2-vs-Rest}~\cite{DMJWBGMY21} & 82,583 & 28,626 & 3,181 & 50,776 \\
& \textsc{7-vs-Rest}~\cite{DMJWBGMY21} & 82,583 & 63,001 & 7,001 & 12,581 \\
& \textsc{Low-vs-High}~\cite{DMJWBGMY21} & 82,583 & 42,791 & 4,755 & 35,037 \\
\hline
GB1~\cite{WDOLS16} 
& \textsc{2-vs-Rest}~\cite{DMJWBGMY21} & 8,733 & 381 & 43 & 8,309 \\
& \textsc{3-vs-Rest}~\cite{DMJWBGMY21} & 8,733 & 2,691 & 299 & 5,743 \\
& \textsc{Low-vs-High}~\cite{DMJWBGMY21} & 8,733 & 4,580 & 509 & 3,644 \\
\hline
Fluorescence~\cite{SBMUMSIBBS16} 
& \textsc{2-vs-Rest} & 54,025 & 12,712 & 1,413 & 39,900 \\
& \textsc{3-vs-Rest}~\cite{XZLZZMLT22} & 54,025 & 21,446 & 5,362 & 27,217 \\
& \textsc{Low-vs-High} & 54,025 & 44,082 & 4,899 & 5,044 \\
\hline
\hline
\end{tabular}
\end{table}

%% file: tables/tbl-results.tex
\begin{table}[!ht]
\setlength\aboverulesep{0pt}
\setlength\belowrulesep{0pt}
\caption{
    Quality in terms of Spearman's $\rho$ correlation with target value.
    NA indicates a non-applicable setting.
    * Used as a feature extractor with pre-trained weights frozen.
    $\dagger$ Results reported in \cite{DMJWBGMY21,XZLZZMLT22}.
    Top-2 performances of each split are marked as \textbf{bold} and \underline{underline}. 
}
\label{table:results}
\setlength{\tabcolsep}{3pt} 
\centering
\small
\resizebox{1.\linewidth}{!}{
\begin{tabular}{l|l|ccc|ccc|ccc}
\toprule
\multirow{3}{*}{\textbf{Category}} & \multirow{3}{*}{\textbf{Model}} & \multicolumn{9}{c}{\textbf{Dataset}} \\
 & & \multicolumn{3}{c|}{\textbf{\AAV}} & \multicolumn{3}{c|}{\textbf{\GB}} &  \multicolumn{3}{c}{\textbf{\Fluorescence} } \\
 & & \textsc{2-vs-R.} & \textsc{7-vs-R.} & \textsc{L.-vs-H.}
 & \textsc{2-vs-R.} & \textsc{3-vs-R.} & \textsc{L.-vs-H.}  
 & \textsc{2-vs-R.} & \textsc{3-vs-R.}  &\textsc{L.-vs-H.} \\
\midrule
\multirow{4}{*}{\textbf{\specialcell{Feature \\ Engineer}}} 
    & Levenshtein
    & 0.578 & 0.550 & 0.251 
    & 0.156 & -0.069 & -0.108 
    & 0.466 & 0.054 & 0.011 \\
    & BLOSUM62
    & NA & NA & NA 
    & 0.128 & 0.005 & -0.127 
    & NA & NA & NA \\
    & DDE
    & 0.649$^\dagger$ & 0.636 & 0.158
    & 0.445$^\dagger$ & 0.816 & 0.306 
    & 0.690 & 0.638$^\dagger$ & 0.159 \\
    & Moran
    & 0.437$^\dagger$ & 0.398 & 0.069
    & 0.069$^\dagger$ & 0.589 & 0.193 
    & 0.445 & 0.400$^\dagger$ & 0.046 \\
\midrule
\multirow{4}{*}{\textbf{\specialcell{Protein Seq. \\ Encoder}}} 
    & LSTM
    & 0.125$^\dagger$ & 0.608 & 0.308
    & -0.002$^\dagger$ & -0.002 & -0.007 
    & 0.256 & 0.494$^\dagger$ & 0.207 \\
    & Transformer
    & 0.681$^\dagger$ & 0.748 & 0.304
    & 0.271$^\dagger$ & 0.877 & 0.474 
    & 0.250 & 0.643$^\dagger$ & 0.161 \\
    & CNN
    & 0.746$^\dagger$ & 0.730 & \underline{0.406}
    & 0.502$^\dagger$ & 0.857 & 0.515 
    & \underline{0.805} & \underline{0.682$^\dagger$} & 0.249 \\
    & ResNet
    & 0.739$^\dagger$ & 0.733 & 0.223
    & 0.133$^\dagger$ & 0.542 & 0.396 
    & 0.594 & 0.636$^\dagger$ & 0.243 \\
\midrule
\multirow{6}{*}{\textbf{\specialcell{Pre-trained \\ PLM}}} 
    & ProtBert
    & 0.794$^\dagger$ & 0.719 & 0.322
    & 0.634$^\dagger$ & 0.866 & 0.308 
    & 0.451 & 0.679$^\dagger$ & 0.201 \\
    & ProtBert*
    & 0.209$^\dagger$ & 0.507 & 0.277
    & 0.123$^\dagger$ & 0.619 & 0.164 
    & 0.403 & 0.339$^\dagger$ & 0.161 \\
    & ESM-1b
    & 0.821$^\dagger$ & 0.735 & 0.385
    & 0.704$^\dagger$ & 0.878 & 0.386 
    & 0.804 & 0.679$^\dagger$ & 0.221 \\
    & ESM-1b*
    & 0.454$^\dagger$ & 0.573 & 0.241
    & 0.337$^\dagger$ & 0.605 & 0.178  
    & 0.528 & 0.430$^\dagger$ & 0.091 \\
    & ESM-1v
    & 0.826 & 0.741 & 0.394
    & 0.721 & \underline{0.884} & 0.390
    & 0.804 & \underline{0.682} & \underline{0.251} \\
    & ESM-1v*
    & 0.533 & 0.580 & 0.171
    & 0.359 & 0.632 & 0.180  
    & 0.562 & 0.563 & 0.070 \\
    & ESM-2
    & 0.824 & 0.734 & 0.390
    & 0.712 & 0.874 & 0.372 
    & 0.791 & 0.668 & 0.201 \\
    & ESM-2*
    & 0.475 & 0.581 & 0.199
    & 0.422 & 0.632 & 0.189  
    & 0.501 & 0.511 & 0.084 \\
\midrule
\multirow{4}{*}{\textbf{\specialcell{GNN-based \\ Methods}}}
    & GCN
    & 0.824 & 0.730 & 0.361
    & 0.745 & 0.865 & 0.466  
    & 0.755 & 0.677 & 0.198 \\
    & GAT
    & 0.821 & 0.741 & 0.369
    & \underline{0.757} & 0.873 & 0.508 
    & 0.768 & 0.667 & 0.208 \\
    & GraphTransf.
    & \underline{0.827} & \underline{0.749} & 0.389
    & 0.753 & 0.876 & \underline{0.548} 
    & 0.780 & 0.678 & 0.231 \\
    & NodeFormer
    & \underline{0.827} & 0.741 & 0.393
    & \underline{0.757} & 0.877 & 0.543 
    & 0.794 & 0.677 & 0.213 \\
\midrule
\multirow{1}{*}{\textbf{\specialcell{Ours}}}
    & \model 
    & \textbf{0.835} & \textbf{0.757} & \textbf{0.433}
    & \textbf{0.768} & \textbf{0.889} & \textbf{0.584} 
    & \textbf{0.809} & \textbf{0.684} & \textbf{0.262} \\
\bottomrule
\end{tabular}
}
\end{table}

%% file: tables/tbl-extensions.tex
\begin{table}[!ht]
\caption{
    Results on \GB datasets (metric: Spearman's $\rho$) of our proposed methods, with different residue embeddings. 
    Top-2 performances of each split marked as \textbf{bold} and \underline{underline}. 
}
\label{table:results_gb1_variants}
\setlength{\tabcolsep}{3pt} 
\centering
\small
\begin{tabular}{l|ccc}
\hline
\hline
\multirow{2}{*}{\textbf{Model}} & \multicolumn{3}{c}{\textbf{Split}} \\
& \textsc{2-vs-R.} & \textsc{3-vs-R.} & \textsc{L.-vs-H.} \\
\hline
State-of-the-art 
& 0.757 & 0.878 & 0.548 \\
\hline
\model ($\Phi^{\textsc{OH}}$) 
& 0.766 & 0.877 & 0.553 \\
\gnn ($\Phi^{\textsc{OH}}$) 
& 0.764 & 0.866 & 0.536 \\
\former ($\Phi^{\textsc{OH}}$) 
& 0.764 & 0.868 & 0.537 \\
\hline
\model ($\Phi^{\textsc{PLM}}$)
& \textbf{0.768} & \textbf{0.881} & \textbf{0.584} \\
\gnn ($\Phi^{\textsc{PLM}}$)
& \underline{0.767} & \underline{0.879} & \underline{0.581} \\
\former ($\Phi^{\textsc{PLM}}$)
& 0.766 & 0.879 & 0.575 \\
\hline
\hline
\end{tabular}
\end{table}

%% file: figures/fig-num-mutation.tex
\begin{figure}[!ht]
\centering
\includegraphics[width=.55\linewidth]{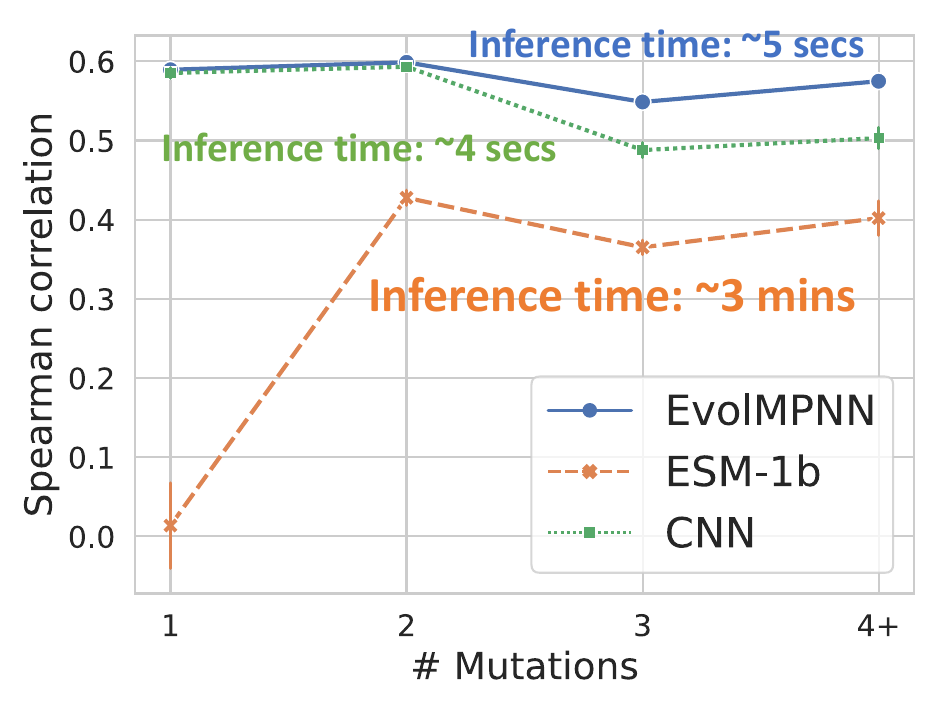}
\caption{
Performance on protein groups of different numbers of mutations, with the \textsc{Low-vs-High} split and avg. epoch inference time on \GB dataset.
}
\label{fig:num_mutations}
\end{figure}

%% file: figures/fig-hyper-param.tex
\begin{figure}[!ht]
\centering
\includegraphics[width=1.\linewidth]{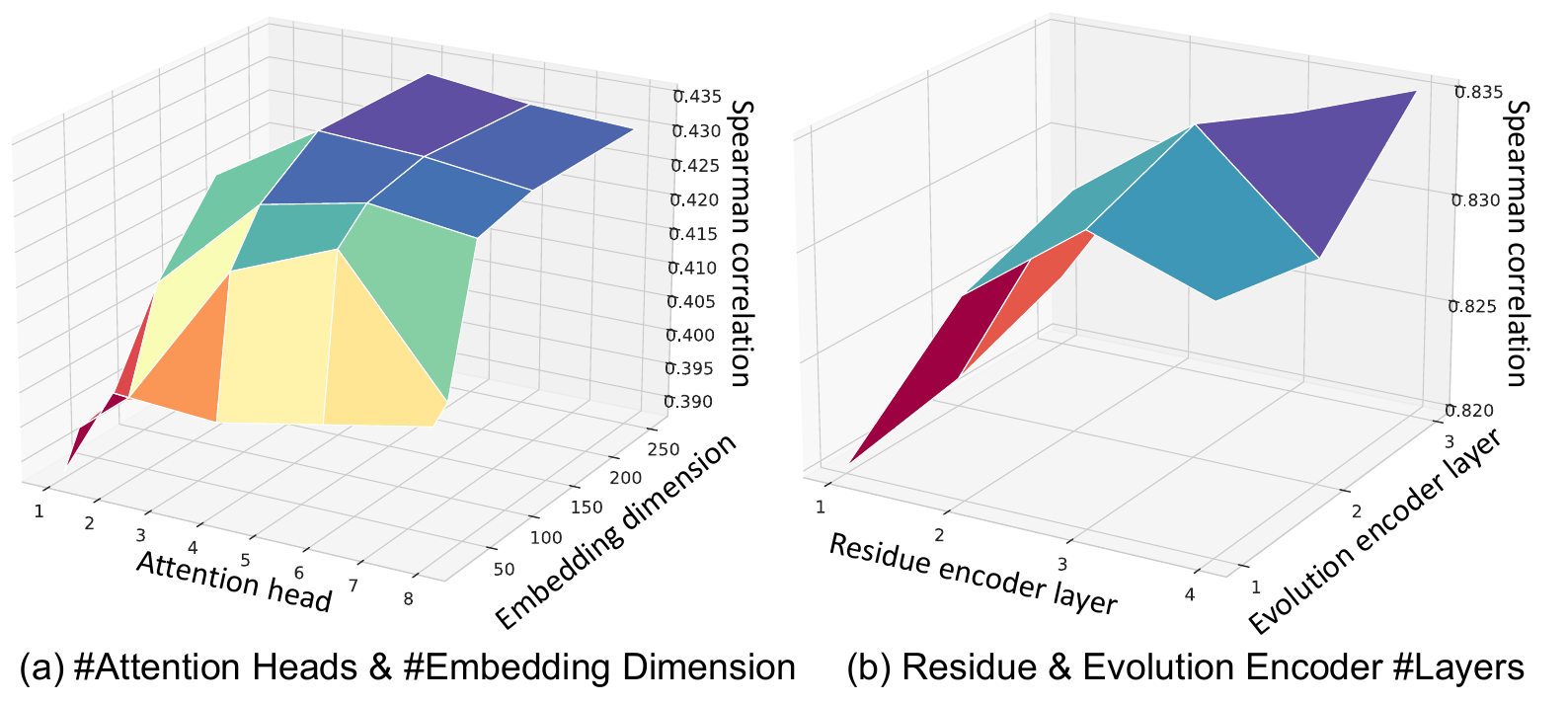}
\caption{
\model performance on \AAV's \textsc{Low-vs-High} (a) and \textsc{2-vs-Rest} (b) splits, with different hyper-parameters.  
}
\label{fig:hyper_param}
\end{figure}

%% file: pages/conclusion.tex
We propose \model that integrates both protein sequence information and evolution information by means of residues to predict the mutational effect on homologous proteins. 
Empirical and theoretical studies show that \model and its extended variants (\gnn and \former) achieve outstanding performance on several benchmark datasets while retaining reasonable computation complexity. 
In future work, we intend to incorporate 3D protein structure information towards general-purpose homologous protein models. 
In addition, it would be interesting to experiment with different approaches to selecting anchor sets, for example, using central nodes-based selections.

%% file: pages/appendix.tex
\section{Theoretical Analysis}
\label{sec:appendix_theoretical_analysis}

Inspired by~\cite{YYL19}, we adopt Bourgain's Theorem~\cite{B85} to guide the random anchor number ($k$) of the evolution encoding layer, such that the resulting embeddings are guaranteed to have low distortion. 
Specifically, distortion measures the faithfulness of the embeddings in preserving distances (in our case, is the differences between protein sequences on a homology network) when mapping from one metric space to another metric space, which can be defined as:
\begin{definition}[Distortion]
\label{def:destortion}
    Given two metric space $(\mathcal{M}, \mathrm{F}_{\textsc{Dist}})$ and $(\mathcal{Z}, \mathrm{F}'_{\textsc{Dist}})$ and a function $f: \mathcal{M} \rightarrow \mathcal{Z}$, $f$ is said to have distortion $\alpha$, if $\forall \mathcal{P}_i, \mathcal{P}_j \in \mathcal{M}$, $\frac{1}{\alpha} \mathrm{F}_{\textsc{Dist}}(\mathcal{P}_i, \mathcal{P}_j) \leq \mathrm{F}'_{\textsc{Dist}} (f(\mathcal{P}_i), f(\textsc{P}_j)) \leq \mathrm{F}_{\textsc{Dist}}(\mathcal{P}_i, \mathcal{P}_j)$. 
\end{definition}

\begin{theorem}[Bourgain Theorem]
\label{th:bourgain}
    Given any finite metric space $(\mathcal{M}, \mathrm{F}_{\textsc{Dist}})$, with $\mid \mathcal{M} \mid = M$, there exists an embedding of $(\mathcal{M}, \mathrm{F}_{\textsc{Dist}})$ into $\mathbb{R}^k$ under any $l_p$ metric, where $k = O(\log^2 M)$, and the distortion of the embedding is $O(\log M)$. 
\end{theorem}

Theorem \ref{th:bourgain} states the Bourgain Theorem~\cite{B85}, which shows the existence of a low distortion embedding that maps from any metric space to the $l_p$ metric space. 

\begin{theorem}[Constructive Proof of Bourgain Theorem]
\label{th:constructive_proof}
    For metric space $(\mathcal{M}, \mathrm{F}_{\textsc{Dist}})$, given $k = \log^2M$ 
    random sets $\{ S_j \}_{j = 1, 2, \dots, \log^2 M} \subset \mathcal{M}$, $S_j$ is chosen by including each point in $\mathcal{M}$ independently with probability $\frac{1}{2^j}$. 
    An embedding method for $\mathcal{P}_i \in \mathcal{M}$ is defined as:
    \begin{equation}
        f(\mathcal{P}_i) = (\frac{\mathrm{F}_{\textsc{Dist}}(\mathcal{P}_i, S_1)}{k}, \frac{\mathrm{F}_{\textsc{Dist}}(\mathcal{P}_i, S_2)}{k}, \dots, \frac{\mathrm{F}_{\textsc{Dist}}(\mathcal{P}_i, S_{\log^2 M})}{k}),
    \end{equation}
    Then, $f$ is an embedding method that satisfies Theorem~\ref{th:bourgain}. 
\end{theorem}

\smallskip\noindent
\textbf{Anchor Selection.}
\model can be viewed as a generalisation of the embedding method of Theorem~\ref{th:constructive_proof}~\cite{LLR95}, where $\mathrm{F}_{\textsc{Dist}} (\cdot)$ is generalised via message passing functions (Eq~\ref{eq:combine_diff}-Eq.~\ref{eq:concat}). 
Therefore, Theorem~\ref{th:constructive_proof} offers a theoretical guide that $O(\log^2M)$ anchors are needed to guarantee low distortion embedding. 
Following this principle, \model choose $k = \log^2M$ random anchors, denoted as $\{ S_{j} \}_{j = 1, 2, \dots, \log^2 M}$, and we sample each protein in $\mathcal{M}$ independently with probability $\frac{1}{2^j}$.